\documentclass[11pt,a4paper]{article}
\usepackage[hyperref]{emnlp2020}
\usepackage{times}
\usepackage{url}
\usepackage{latexsym}

\usepackage{pgfplots}
\pgfplotsset{compat=1.14} 
\usepackage{graphicx}
\usepackage{xspace}
\usepackage{url}
\usepackage[shortlabels]{enumitem}
\usepackage{pifont}
\usepackage{amsmath}
\usepackage{textcomp}
\usepackage[font=small,labelfont=bf]{caption}

\newcommand{\bert}{\textsc{bert}\xspace}
\newcommand{\legalbert}{\textsc{legal-bert}\xspace}
\newcommand{\legalberta}{\textsc{legal-bert-fp}\xspace}
\newcommand{\legalbertp}{\textsc{legal-bert-sc}\xspace}

\newcommand{\legalbertsmall}{\textsc{legal-bert-small}\xspace}
\newcommand{\bertbase}{\textsc{bert-base}\xspace}
\newcommand{\eurlexdata}{\textsc{eurlex57k}\xspace}
\newcommand{\echrdata}{\textsc{echr-cases}\xspace}
\newcommand{\contractsdata}{\textsc{contracts-ner}\xspace}
\newcommand{\glue}{\textsc{glue}\xspace}

\newcommand{\squad}{\textsc{squad}\xspace}
\newcommand{\race}{\textsc{race}\xspace}
\newcommand{\nlp}{\textsc{nlp}\xspace}
\newcommand{\eurlex}{\textsc{eurlex}\xspace}
\newcommand{\eu}{\textsc{eu}\xspace}
\newcommand{\ecj}{\textsc{ecj}\xspace}
\newcommand{\uk}{\textsc{uk}\xspace}
\newcommand{\echr}{\textsc{echr}\xspace}
\newcommand{\hudoc}{\textsc{hudoc}\xspace}
\newcommand{\us}{\textsc{us}\xspace}

\newcommand{\edgar}{\textsc{edgar}\xspace}

\newcommand{\eurovoc}{\textsc{eurovoc}\xspace}
\newcommand{\biobert}{\textsc{biobert}\xspace}

\newcommand{\scibert}{\textsc{scibert}\xspace}

\usepackage{graphicx}
\usepackage{microtype}

\aclfinalcopy

\title{LEGAL-BERT: The Muppets straight out of Law School}

\author{Ilias Chalkidis$^{\;\dagger\;\ddagger}$ \qquad Manos Fergadiotis$^{\;\dagger\;\ddagger}$ \\ \textbf{Prodromos Malakasiotis$^{\;\dagger\;\ddagger}$} \qquad \textbf{Nikolaos Aletras$\;^{*}$} \qquad \textbf{Ion Androutsopoulos$^{\;\dagger\;\ddagger}$} \\$^{\dagger\;}$Department of Informatics, Athens University of Economics and Business \\ $^{\ddagger\;}$Institute of Informatics \& Telecommunications, NCSR ``Demokritos'' \\ $^{*\;}$Computer Science Department, University of Sheffield, UK \\ {\tt [ihalk,fergadiotis,rulller,ion]@aueb.gr} \\ {\tt n.aletras@sheffield.ac.uk}}

\date{}

\begin{document}
\maketitle
\begin{abstract}
 
\bert has achieved impressive performance in several \nlp tasks. However, there has been limited investigation on its adaptation guidelines in specialised domains. Here we focus on the legal domain, where we explore several approaches for applying \bert models to downstream legal tasks, evaluating on multiple datasets. Our findings indicate that the previous guidelines for pre-training and fine-tuning, often blindly followed, do not always generalize well in the legal domain. Thus we propose a systematic investigation of the available strategies when applying \bert in specialised domains. These are: (a) use the original \bert out of the box, (b) adapt \bert by additional pre-training on domain-specific corpora, and (c) pre-train \bert from scratch on domain-specific corpora. We also propose a broader hyper-parameter search space when fine-tuning for downstream tasks and we release \legalbert, a family of \bert models intended to assist legal \nlp research, computational law, and legal technology applications.
\end{abstract}

\begin{table*}[ht!]
    \centering
    \resizebox{\textwidth}{!}{
    \begin{tabular}{l|c|c|l}
         Corpus & No. documents & Total Size in GB & Repository  \\
         \hline
        \eu legislation & 61,826 & 1.9 (16.5\%) & \eurlex (\url{eur-lex.europa.eu}) \\
        \uk legislation & 19,867 & 1.4  (12.2\%) & \textsc{legislation.gov.uk} (\url{http://www.legislation.gov.uk}) \\
        European Court of Justice (\ecj) cases & 19,867 & 0.6 (\;\;5.2\%) & \eurlex (\url{eur-lex.europa.eu}) \\
        European Court of Human Rights (\echr) cases &  12,554 & 0.5 (\;\;4.3\%) & \hudoc (\url{http://hudoc.echr.coe.int}) \\
        \us court cases & 164,141 & 3.2 (27.8\%) & \textsc{case law access project} (\url{https://case.law})\\
        \us contracts & 76,366 & 3.9 (34.0\%) & \textsc{sec-edgar} (\url{https://www.sec.gov/edgar.shtml})
    \end{tabular}
    }
    \caption{Details on the training corpora used to pre-train the different variations of \legalbert. All repositories have open access, except from the Case Law Access Project, where access is granted to researchers upon request.}
    \label{tab:data}
\end{table*}

\section{Introduction}
Pre-trained language models based on Transformers~\cite{Vaswani2017}, such as \bert~\cite{devlin2019} and its variants \cite{roberta,xlnet,albert}, have achieved state-of-the-art results in several downstream \nlp tasks on generic benchmark datasets, such as \glue~\cite{wang2018glue}, \squad~\cite{rajpurkar2016squad}, and \race~\cite{lai2017race}.

\begin{figure}[t!]
    \centering
    \includegraphics[width=\columnwidth]{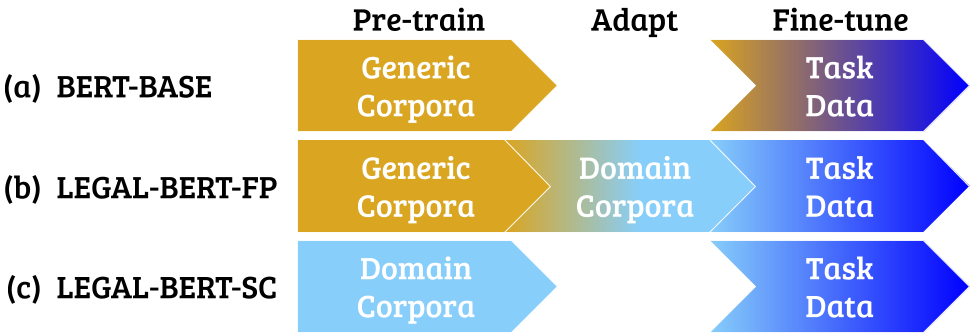}
    \caption{The three alternatives when employing \bert for \nlp tasks in specialised domains: (a) use \bert out of the box, (b) further pre-train \bert (\texttt{FP}), and (c) pre-train \bert from scratch (\texttt{SC}). All strategies have a final fine-tuning step.}
    \label{fig:bert_adapt}
    \vspace{-6mm}
\end{figure}

Typically, transfer learning with language models requires a computationally heavy step where the language model is pre-trained on a large corpus and a less expensive step where the model is fine-tuned for downstream tasks. When using \bert, the first step can be omitted as the pre-trained models are publicly available. Being pre-trained on generic corpora (e.g., Wikipedia, Children's  Books, etc.) \bert has been reported to under-perform in specialised domains, such as biomedical or scientific text \cite{lee2019,beltagy2019}. To overcome this limitation there are two possible strategies; either further pre-train (\texttt{FP}) \bert on domain specific corpora, or pre-train \bert from scratch (\texttt{SC}) on domain specific corpora. Consequently, to employ \bert in specialised domains one may consider three alternative strategies before fine-tuning for the downstream task (Figure~\ref{fig:bert_adapt}): (a) use \bert out of the box, (b) further pre-train (\texttt{FP}) \bert on domain-specific corpora, and (c) pre-train \bert from scratch (\texttt{SC}) on domain specific corpora with a new vocabulary of sub-word units.

In this paper, we systematically explore strategies (a)--(c) in the legal domain, where \bert adaptation has yet to be explored. As with other specialised domains, legal text (e.g., laws, court pleadings, contracts) has distinct characteristics compared to generic corpora, such as specialised vocabulary, particularly formal syntax, semantics based on extensive domain-specific knowledge etc., to the extent that legal language is often classified as a `sublanguage' \cite{Tiersma1999,Williams2007,Haigh2018}. Note, however, that our work contributes more broadly towards a better understanding of domain adaptation for specialised domains. Our key findings are: 
(i) Further pre-training (\texttt{FP}) or pre-training \bert from scratch (\texttt{SC}) on domain-specific corpora, performs better than using \bert out of the box for domain-specific tasks; both strategies are mostly comparable in three legal datasets.
(ii) Exploring a broader hyper-parameter range, compared to the guidelines of \newcite{devlin2019}, can lead to substantially better performance.
(iii) Smaller \bert-based models can be competitive to larger, computationally heavier ones in specialised domains.
Most importantly, (iv) we release \legalbert, a family of \bert models for the legal domain, intended to assist legal \nlp research, computational law, and legal technology applications.\footnote{All models and code examples are available at: \url{https://huggingface.co/nlpaueb}.}
This family includes \legalbertsmall, a light-weight model pre-trained from scratch on legal data, which achieves comparable performance to larger models, while being much more efficient (approximately 4 times faster) with a smaller environmental footprint \cite{strubell2019}.

\section{Related Work}

Most previous work on the  domain-adaptation of \bert and variants does not systematically explore the full range of the above strategies and mainly targets the biomedical or broader scientific domains. \newcite{lee2019} studied the effect of further pre-training \bertbase on biomedical articles for 470k steps.
The resulting model (\biobert) was evaluated on biomedical datasets, reporting performance improvements compared to \bertbase. Increasing the additional domain-specific pre-training to 1M steps, however, did not lead to any clear further improvements. \newcite{alsentzer2019} released Clinical \bert and Clinical \biobert by further pre-training \bertbase and \biobert, respectively, on clinical notes for 150k steps. Both models were reported to outperform \bertbase. 
In other related work, \newcite{beltagy2019} released \scibert, a family of \bert-based models for scientific text, with emphasis on the biomedical domain. Their models were obtained either by further pre-training (\texttt{FP}) \bertbase, or by pre-training \bertbase from scratch (\texttt{SC}) on a domain-specific corpus, i.e., the model is randomly initialized and the vocabulary was created from scratch.
Improvements were reported in downstream tasks in both cases. \newcite{sung2019} further pre-trained \bertbase on textbooks and question-answer pairs to improve short answer grading for intelligent tutoring systems. 

One shortcoming is that all previous work does not investigate the effect of varying the number of pre-training steps, with the exception of \newcite{lee2019}. More importantly, when fine-tuning for the downstream task, all previous work blindly adopts the hyper-parameter selection guidelines of \newcite{devlin2019} without further investigation.
Finally, no previous work considers the effectiveness and efficiency of smaller models (e.g., fewer layers) in specialised domains. The full capacity of larger and computationally more expensive models may be unnecessary in specialised domains, where syntax may be more standardized, the range of topics discussed may be narrower, terms may have fewer senses etc.  
We also note that although \bert is the current state-of-the-art in many legal \nlp tasks \cite{chalkidis2019, chalkidis2019b,chalkidis2019d}, no previous work has considered its adaptation for the legal domain.

\section{LEGAL-BERT: A new family of BERT models for the legal domain}
\label{sec:legalbert}

\noindent\textbf{Training corpora:} To pre-train the different variations of \legalbert, we collected 12 GB of diverse English legal text from several fields (e.g., legislation, court cases,  contracts) scraped from publicly available resources (see Table \ref{tab:data}).\vspace{1mm}

\noindent\textbf{\legalberta:}  Following \newcite{devlin2019}, we run additional pre-training steps of \bertbase on domain-specific corpora. While \newcite{devlin2019} suggested additional steps up to 100k, we also pre-train models up to 500k to examine the effect of prolonged in-domain pre-training when fine-tuning on downstream tasks. 
\bertbase has been pre-trained for significantly more steps in generic corpora (e.g., Wikipedia, Children's  Books), thus it is highly skewed towards generic language, using a  vocabulary of 30k sub-words that better suits these generic corpora. Nonetheless we expect that prolonged in-domain pre-training will be beneficial.\vspace{1mm}

\noindent\textbf{\legalbertp} has the same architecture as \bertbase with 12 layers, 768 hidden units and 12 attention heads (110M parameters). We use this architecture in all our experiments unless otherwise stated. We use a newly created vocabulary of equal size to \bert's vocabulary.\footnote{We use Google's sentencepiece library (\url{https://github.com/google/sentencepiece}.)} We also experiment with \legalbertsmall, a substantially smaller model, with 6 layers, 512 hidden units, and 8 attention heads (35M parameters, 32\% the size of \bertbase). This light-weight model, trains approx.\ 4 times faster, while also requiring fewer hardware resources.\footnote{Consult Appendix C for a comparison on hardware resources as well as training and inference times.} Our hypothesis is that such a specialised \bert model can perform well against generic \bert models, despite its fewer parameters.\vspace{1mm}

\section{Experimental Setup}
\label{sec:experimentalsetup}

\begin{figure}[t!]
    \centering
    \includegraphics[width=\columnwidth]{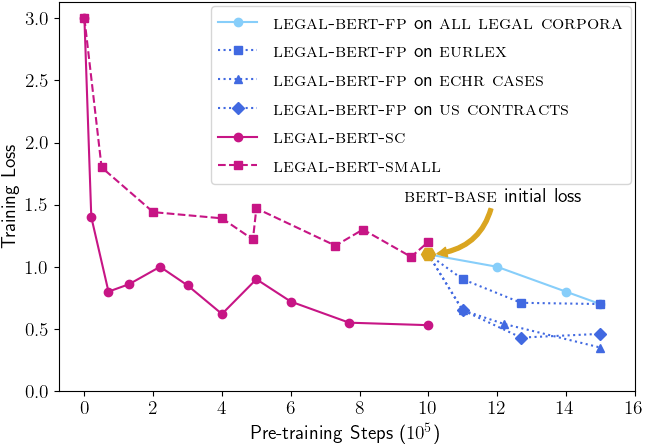}
    \caption{Train losses for all \legalbert versions.}
    \label{fig:trainloss}
\end{figure}

\begin{figure*}[t!]
\begin{center}
\includegraphics[width=\textwidth]{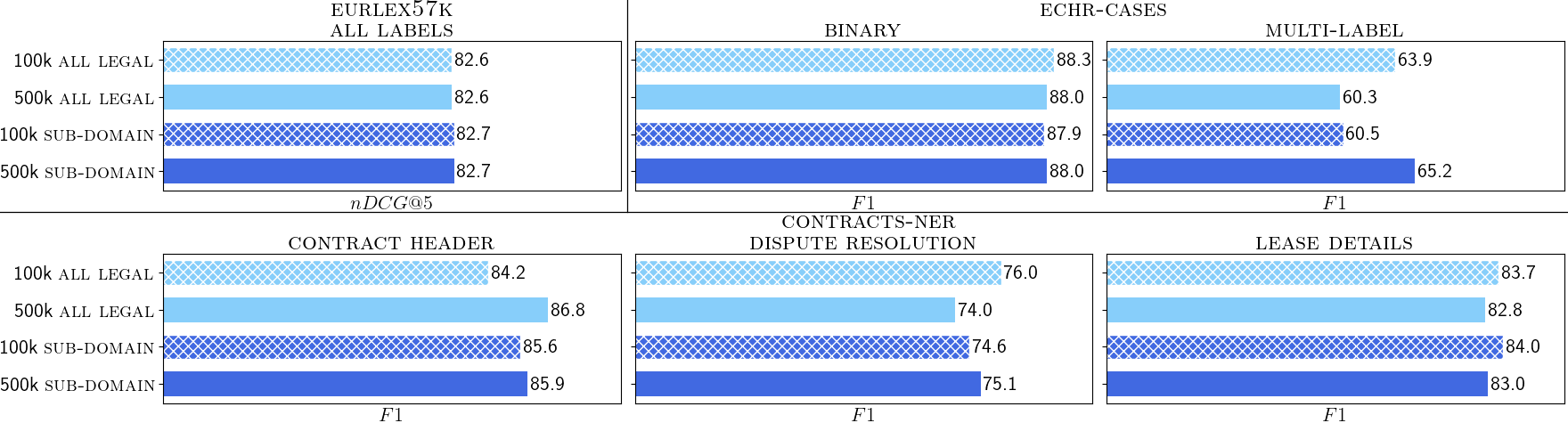}
\end{center}
\caption{End-task results on development data across all datasets for \legalberta variants.}
\label{fig:bar_chart_fp}
\end{figure*}

\begin{figure*}[t!]
\begin{center}
\includegraphics[width=\textwidth]{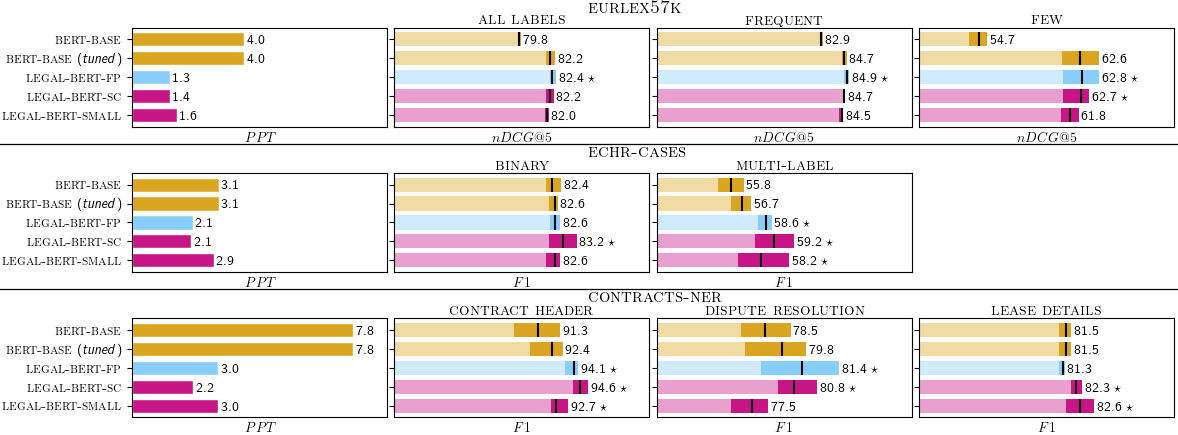}
\end{center}
\caption{Perplexities ({\small \textit{PPT}}) and end-task results on test data across all datasets and all models considered. The reported results are averages over multiple runs also indicated by a vertical black line in each bar. The transparent and opaque parts of each bar show the minimum and maximum scores of the runs, respectively. A star indicates versions of \legalbert that perform better on average than the tuned \bertbase.}
\label{fig:bar_chart}
\end{figure*}


\noindent\textbf{Pre-training Details:} To be comparable with \bert, we train \legalbert for 1M steps (approx.\ 40 epochs) over all corpora (Section \ref{sec:legalbert}), in batches of 256 samples, including up to 512 sentencepiece tokens. We used Adam with learning rate of $1\mathrm{e-}4$, as in the original implementation. We trained all models with the official \bert code\footnote{ \url{github.com/google-research/bert}} using v3 \textsc{tpu}s with 8 cores from Google Cloud Compute Services.\vspace{1mm}

\noindent\textbf{Legal \nlp Tasks:} We evaluate our models on text classification and sequence tagging using three datasets. \eurlexdata~\cite{chalkidis2019c} is a large-scale multi-label text classification dataset of \eu laws, also suitable for few and zero-shot learning. \echrdata~\cite{chalkidis2019b} contains cases from the European Court of Human Rights~\cite{aletras2016predicting} and can be used for binary and multi-label text classification. Finally, \contractsdata~\cite{chalkidis2017,chalkidis2019d} is a dataset for named entity recognition on \us contracts consisting of three subsets, \emph{contract header}, \emph{dispute resolution}, and \emph{lease details}. 
We replicate the experiments of~\newcite{chalkidis2019,chalkidis2019b,chalkidis2019d} when fine-tuning \bert for all datasets.\footnote{For implementation details, see Appendices A and B.}
\vspace{1mm}

\noindent\textbf{Tune your Muppets!} As a rule of thumb to fine-tune \bert for downstream tasks, \newcite{devlin2019} suggested a minimal hyper-parameter tuning strategy relying on a grid-search on the following ranges: learning rate $\in\{2\mathrm{e-}5, 3\mathrm{e-}5, 4\mathrm{e-}5, 5\mathrm{e-}5\}$, number of training epochs $\in\{3,4\}$, batch size $\in\{16,32\}$ and fixed dropout rate of $0.1$. These not well justified suggestions are blindly followed in the literature \cite{lee2019, alsentzer2019, beltagy2019, sung2019}. Given the relatively small size of the datasets, we use batch sizes $\in\{4,8,16,32\}$.  Interestingly, in preliminary experiments, we found that some models still underfit after 4 epochs, the maximum suggested, thus we use early stopping based on validation loss, without a fixed maximum number of training epochs. We also consider an additional lower learning rate ($1\mathrm{e-}5$) to avoid overshooting local minima, and an additional higher drop-out rate ($0.2$) to improve regularization. Figure~\ref{fig:bar_chart} (top two bars) shows that our enriched grid-search (\emph{tuned}) has a substantial impact in most of the end-tasks compared to the default hyper-parameter strategy of \newcite{devlin2019}.\footnote{In the \emph{lease details} subset of \contractsdata, the optimal hyper-parameters fall in the ranges of \newcite{devlin2019}.} We adopt this strategy for \legalbert.\vspace{1mm}

\section{Experimental Results}
\noindent\textbf{Pre-training Results:}
Figure~\ref{fig:trainloss} presents the training loss across pre-training steps for all versions of \legalbert. \legalbertp performs much better on the pre-training objectives than \legalbertsmall, which was highly expected, given the different sizes of the two models. At the end of its pre-training, \legalbertsmall has similar loss to that of \bertbase pre-trained on generic corpora (arrow in Figure~\ref{fig:trainloss}). When we consider the additional pre-training of \bert on legal corpora (\legalberta), we observe that it adapts faster and better in specific sub-domains (esp.\ \echr cases, \us contracts), comparing to using the full collection of legal corpora, where the training loss does not reach that of \legalbertp.\vspace{1mm}

\noindent\textbf{End-task Results:}
Figure~\ref{fig:bar_chart_fp} presents the results of all \legalberta variants on development data. The optimal strategy for further pre-training varies across datasets. Thus in subsequent experiments on test data, we keep for each end-task the variant of \legalberta with the best development results.

Figure~\ref{fig:bar_chart} shows the perplexities and end-task results (minimum, maximum, and averages over multiple runs) of all \bert variants considered, now on test data. Perplexity indicates to what extent a \bert variant predicts the language of an end-task. We expect models with similar perplexities to also have similar performance. In all three datasets, a \legalbert variant almost always leads to better results than the tuned \bertbase. In \eurlexdata, the improvements are less substantial for \emph{all}, \emph{frequent}, and \emph{few} labels (0.2\%), also in agreement with the small drop in perplexity (2.7). In \echrdata, we again observe small differences in perplexities (1.1 drop) and in the performance on the binary classification task (0.8\% improvement). On the contrary, we observe a more substantial improvement in the more difficult multi-label task (2.5\%) indicating that the \legalbert variations benefit from in-domain knowledge. On \contractsdata, the drop in perplexity is larger (5.6), which is reflected in the increase in $F1$ on the \emph{contract header} (1.8\%) and \emph{dispute resolution} (1.6\%) subsets. In the \emph{lease details} subset, we also observe an improvement (1.1\%). Impressively, \legalbertsmall is comparable to \legalbert across most datasets, while it can fit in most modern \textsc{gpu} cards. This is important for researchers and practitioners with limited access to large computational resources. It also provides a more memory-friendly basis for more complex \bert-based architectures. For example, deploying a hierarchical version of \bert for \echrdata \citep{chalkidis2019b} leads to a $4\times$ memory increase.

\section{Conclusions and Future Work}
We showed that the best strategy to port \bert to a new domain may vary, and one may consider either further pre-training or pre-training from scratch. Thus, we release \legalbert, a family of \bert models for the legal domain achieving state-of-art results in three end-tasks. 
Notably, the performance gains are stronger in the most challenging end-tasks (i.e., \emph{multi-label} classification in \echrdata and \emph{contract header}, \emph{lease details} in \contractsdata) where in-domain knowledge is more important. 
We also release \legalbertsmall, which is 3 times smaller but highly competitive to the other versions of \legalbert. Thus, it can be adopted more easily in low-resource test-beds. 
Finally, we show that an expanded grid search when fine-tuning \bert for end-tasks has a drastic impact on performance and thus should always be adopted. In future work, we plan to explore the performance of \legalbert in more legal datasets and tasks. We also intend to explore the impact of further pre-training \legalbertp and \legalbertsmall on specific legal sub-domains (e.g., \eu legislation).

\section*{Acknowledgments}
This project was supported by the Google Cloud Compute (\textsc{gcp}) research program,  while we also used a Google Cloud \textsc{tpu} v3-8 for free provided by the TensorFlow Research Cloud (\textsc{tfrc}) program\footnote{\url{https://www.tensorflow.org/tfrc}}. We are grateful to both Google programs.

\bibliographystyle{acl_natbib}
\bibliography{emnlp2020}

\appendix

\section{Legal NLP datasets}
\label{app:task_data}
Bellow are the details of the legal \nlp datasets we used for the evaluation of our models: 

\begin{itemize}
    \item \eurlexdata \cite{chalkidis2019c} contains 57k legislative documents from \eurlex with an average length of 727 words. All documents have been annotated by the Publications Office of \eu with concepts from \eurovoc.\footnote{\url{http://eurovoc.europa.eu/}} The average number of labels per document is approx.\ 5, while many of them are rare. The dataset is split into \emph{training} (45k),  \emph{development}  (6k),  and  \emph{test} (6k) documents. 
    
    \item \echrdata \cite{chalkidis2019b} contains approx.\ 11.5k cases from \echr's public database. For each case, the dataset provides a list of \emph{facts}. Each case is also mapped to \emph{articles} of the Human Rights Convention that were violated (if any).  The dataset can be used for binary classification, where the task is to identify if there was a violation or not, and for multi-label classification where the task is to identify the violated articles.
    
    \item \contractsdata \cite{chalkidis2017,chalkidis2019d} contains approx.\ 2k \us contracts from \edgar. Each contract has been annotated with multiple contract elements such as \emph{title}, \emph{parties}, \emph{dates of interest}, \emph{governing law}, \emph{jurisdiction}, \emph{amounts} and \emph{locations}, which have been organized in three groups (\emph{contract header}, \emph{dispute resolution}, \emph{lease details}) based on their position in contracts.
\end{itemize}

\section{Implementation details and results on downstream tasks}
\label{appendix:evaluation}

Below we describe the implementation details for fine-tuning \bert and \legalbert on the three downstream tasks:

\begin{description}
\item[\eurlexdata:] We replicate the experiments of \newcite{chalkidis2019}, where a linear layer with $L$ (number of labels) $\mathrm{sigmoid}$ activations was placed on top of \bert's {\tt[CLS]} final representation. We follow the same configuration for all \legalbert variations.

\item[\echrdata:] We replicate the best method of \newcite{chalkidis2019b}, which is a hierarchical version of \bert, where initially a shared \bert encodes each case fact independently and produces $N$ fact embeddings ({\tt[CLS]} representations). A self-attention mechanism, similar to \newcite{Yang2016}, produces the final document representation. A linear layer with $\mathrm{softmax}$ activation gives the final scores.

\item[\contractsdata] We replicate the experiments of \newcite{chalkidis2019d} in all of their three parts (\emph{contract header}, \emph{dispute resolution}, \emph{lease details}). In these experiments, the final representations of the original \bert for all (sentencepiece) tokens in the sequence are fed to a linear \textsc{crf} layer.
\end{description}

We again follow \newcite{chalkidis2019,chalkidis2019b,chalkidis2019d} in the reported evaluation measures.

\section{Efficiency comparison for various \bert-based models}

\begin{table}[ht]
    \centering
    \resizebox{\columnwidth}{!}{%
    \begin{tabular}{l|c|c|c|c|c|c|c|c}
    \hline
    &        &     &      &      &          & \multicolumn{2}{c|}{Training}   & Inference\\
    Model.                    & Params & $T$ & $HU$ & $AH$ & Max $BS$ & \multicolumn{2}{c|}{Speed}      & Speed \\\cline{7-9}
                              &        &     &      &      &          & $BS=1$       & $BS=\max$        & $BS=1$\\
    \hline
    \bertbase                 & 110M   & 12 & 768   & 12   & 6        & $1.00\times$ & $1.00\times$     & $1.00\times$ \\
    \textsc{albert}.          & 12M    & 12 & 768   & 12   & 12       & $1.26\times$ & $1.21\times$     & $1.00\times$ \\
    \textsc{albert-large}     & 18M    & 24 & 1024  & 12   & 4        & $0.49\times$ & $0.37\times$     & $0.36\times$ \\
    \textsc{distil-bert}      & 66M    & 6  & 768   & 12   & 16       & $1.66\times$ & $2.36\times$     & $1.70\times$ \\
    \textsc{legal-bert}       & 110M   & 12 & 768   & 12   & 6        & $1.00\times$ & $1.00\times$     & $1.00\times$ \\
    \textsc{legal-bert-small} & 35M    & 6  & 512   & 8    & 26       & $2.43\times$ & $4.00\times$     & $1.70\times$ \\
    \end{tabular}
}
\caption{Comparison of \bert{-based} models for different batch sizes ($BS$) in a single 11GB \textsc{nvidia-2080Ti}. Resource efficiency of the models mostly relies on the number of hidden units ($HU$), attentions heads ($AH$) and Transformer blocks $T$, rather than the number of parameters.}

\label{tab:gpus}
\end{table}

Recently there has been a debate on the over-parameterization of \bert \cite{kitaev2020, Rogers2020API}. Towards that directions most studies suggest a parameter sharing technique \cite{albert} or distillation of \bert by decreasing the number of layers \cite{distilbert}. However the main bottleneck of transformers in modern hardware is not primarily the total number of parameters, misinterpreted into the number of stacked layers. Instead Out Of Memory (\textsc{oom}) issues mainly happen as a product of wider models in terms of hidden units' dimensionality and the number of attention heads, which affects gradient accumulation in feed-forward and multi-head attention layers (see Table \ref{tab:gpus}). Table~\ref{tab:gpus} shows that \legalbertsmall despite having $3\times$ and $2\times$ the parameters of \textsc{albert} and \textsc{albert-large} has faster training and inference times. We expect models overcoming such limitations to be widely adopted by researchers and practitioners with limited resources. Towards the same direction Google released several lightweight versions of \bert.\footnote{\url{https://github.com/google-research/bert}}

\end{document}